\documentclass[runningheads]{llncs}

\usepackage{siunitx}

\mathchardef\mhyphen="2D
\newcolumntype{C}{@{\extracolsep{6pt}}c@{\extracolsep{3pt}}}%
\newcolumntype{L}{@{\extracolsep{6pt}}l@{\extracolsep{3pt}}}%

\makeatletter
\DeclareRobustCommand\onedot{\futurelet\@let@token\@onedot}

\makeatother
\usepackage{graphicx}
\usepackage{makecell}
\usepackage{booktabs}
\usepackage{multirow}
\usepackage[bottom]{footmisc}
\usepackage{amsfonts,amssymb}
\usepackage{svg}
\usepackage{amsmath}
\usepackage{graphicx,verbatim}
\usepackage{amssymb}
\usepackage{caption}
\usepackage[table]{xcolor}
\usepackage{colortbl}

\newcommand{\pmS}[1]{\scalebox{0.8}{$\pm$\,#1}} 

\newcommand{\pct}[1]{\num[scientific-notation=false,round-mode=places,round-precision=1]{\fpeval{100*(#1)}}}

\definecolor{C2}{RGB}{232,212,232}
\definecolor{C1}{RGB}{251,217,211}
\definecolor{red}{RGB}{176,26,24}
\definecolor{purple}{RGB}{104,52,154}
\definecolor{gray}{RGB}{128,128,128}

\usepackage[T1]{fontenc}
\usepackage{graphicx,verbatim}
\newcommand{\M}{LoGo-MR}
\newcommand{\MV}{LoGo$^{3}$-MR}

\begin{document}
\title{\M: Screening Breast MRI for Cancer Risk Prediction by Efficient Omni-Slice Modeling}
\titlerunning{Screening Breast MRI for Cancer Risk Prediction}

\author{
Xin Wang\inst{1,2}$^{\dagger}$ \and
Yuan Gao\inst{1,3}$^{\dagger}$ \and
George Yiasemis\inst{1} \and
Antonio Portaluri\inst{1,2} \and \\
Zahra Aghdam\inst{1,2} \and
Muzhen He\inst{1} \and
Luyi Han\inst{1,2} \and
Yaofei Duan\inst{1} \and
Chunyao Lu\inst{1,2} \and
Xinglong Liang\inst{1,2} \and
Tianyu Zhang\inst{1,2} \and
Vivien van Veldhuizen\inst{1} \and
Yue Sun\inst{4} \and \\
Tao Tan\inst{4} \and
Ritse Mann\inst{1,2} \and
Jonas Teuwen\inst{1,2}$^{*}$
}

\authorrunning{X. Wang, Y. Gao, et al.}

\institute{
Netherlands Cancer Institute \and
Radboud University Medical Center \and
University Medical Center Utrecht \and
Macao Polytechnic University \\
$^{\dagger}$ These authors contributed equally to this work. \\
$^{*}$ Corresponding author: j.teuwen@nki.nl
}

\maketitle              
\begin{abstract}
  Efficient and explainable breast cancer (BC) risk prediction is critical for large-scale population-based screening. Breast MRI provides functional information for personalized risk assessment. 
  Yet effective modeling remains challenging as fully 3D CNNs capture volumetric context at high computational cost, whereas lightweight 2D CNNs fail to model inter-slice continuity. 
  Importantly, breast MRI modeling for short- and long-term BC risk stratification remains underexplored.
  In this study, we propose \M, a 2.5D local–global structural modeling framework for five-year BC risk prediction. 
  Aligned with clinical interpretation, our framework first employs neighbor-slice encoding to capture subtle local cues linked to short-term risk. It then integrates transformer-enhanced multiple-instance learning (MIL) to model distributed global patterns related to long-term risk and provide interpretable slice importance.
  We further apply this framework across axial, sagittal, and coronal planes as \MV~to capture complementary volumetric information. This multi-plane formulation enables voxel-level risk saliency mapping, which may assist radiologists in localizing risk-relevant regions during breast MRI interpretation. 
  Evaluated on a large breast MRI screening cohort ($\sim$7.5K), our method outperforms 2D/3D baselines and existing SOTA MIL methods, achieving AUCs of 0.77-0.69 for 1- to 5-year prediction and improving C-index by $\sim$6\% over 3D CNNs. \MV~further improves overall performance with interpretable localization across three planes, and validation across seven backbones shows consistent gains. These results highlight the clinical potential of efficient MRI-based BC risk stratification for large-scale screening.
  Code will be released publicly.
  
  \keywords{Breast cancer screening \and MRI \and Risk prediction \and MIL}
\end{abstract}
\section{Introduction}

Population-based breast cancer (BC) screening programs, which invite eligible women to undergo routine examinations, are implemented worldwide to enable earlier detection and reduce BC mortality \cite{yala2021toward}. For women with higher lifetime risk, guidelines escalate screening intensity and often recommend breast dynamic contrast-enhanced (DCE) MRI because of its high sensitivity \cite{yala2021toward}. However, this approach is constrained by cost and limited clinical capacity~\cite{lehman2016screening,mann2020novel}, underscoring the need to reserve MRI for women who could benefit more. Further risk stratification within this high-risk population is therefore critical for tailoring screening pathways and allocating MRI resources efficiently.

\begin{figure}[!b]
\centering
  \includegraphics[width=\textwidth, height=0.265\textheight]{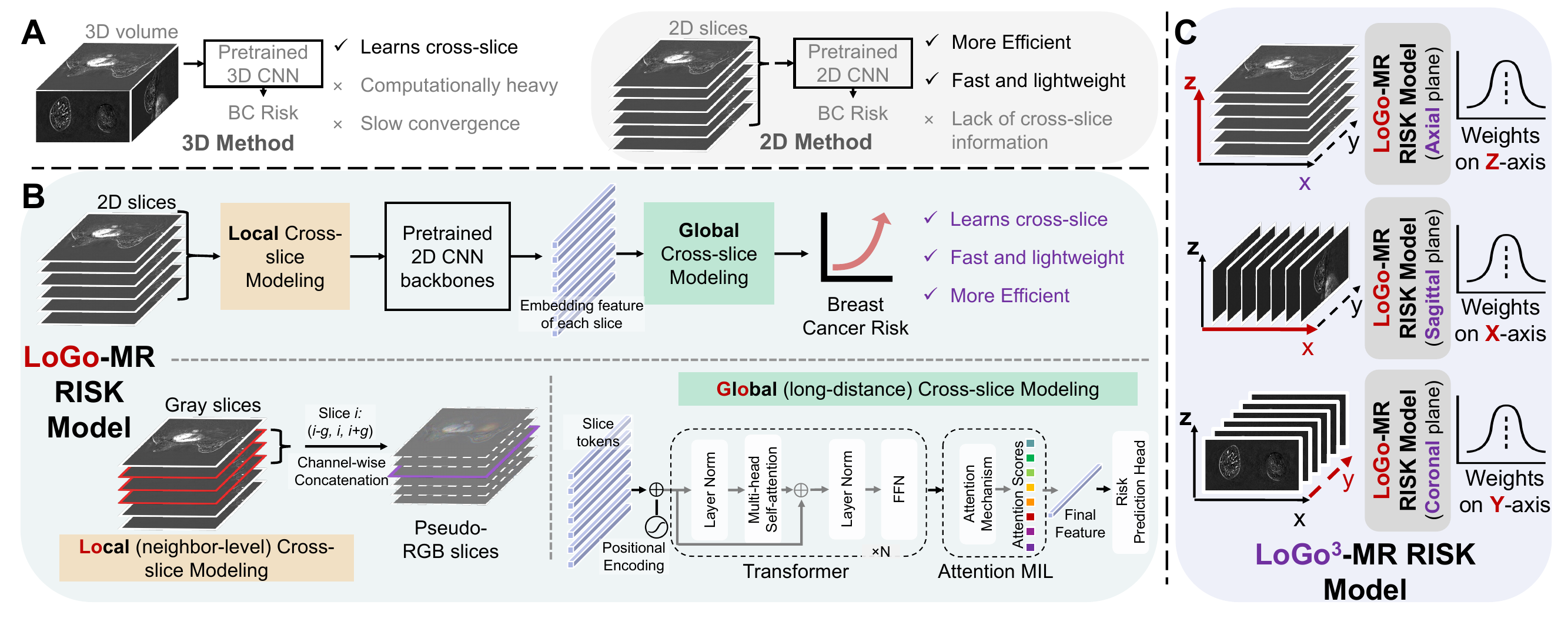}
  \caption{Comparison of 3D, 2D, and the proposed \M~and \MV~frameworks for BC risk prediction. \M~combines local slice fusion with global MIL-based sequence modeling to capture both short-range anatomical continuity and long-range slice dependencies, while providing interpretable slice-level importance. \MV~extends this strategy across axial, sagittal, and coronal planes to capture complementary anatomical cues.
  } \label{fig:method}
\end{figure}

Beyond traditional risk factor-based approaches \cite{tyrer2004breast}, deep learning has enabled risk stratification directly from routinely acquired 2D screening mammograms~\cite{yala2021toward,wang2025predicting,wang2025mammo}. However, for breast MRI, prior work has primarily relied on quantitative MRI-derived biomarkers, such as background parenchymal enhancement (BPE), with modest predictive performance reported~\cite{geissler2025mri,wang2023parenchymal}. 
Modeling MRI is inherently challenging, as clinically relevant cues often span contiguous slices, necessitating volumetric modeling, yet fully 3D CNNs (Fig. \ref{fig:method}~A)
remain computationally intensive and difficult to train efficiently~\cite{wang2025accurate,li20262d,wang20232}. 
In screening settings requiring rapid AI inference, such computational demands may increase operational burden and hinder large-scale deployment.
To balance efficiency and limited contextual modeling, 
recent approaches explored 2D architectures with slice-level encoding (e.g., multi-slice inputs or sparse sampling) \cite{wang20232,pang2025interpretable}.
However, these strategies leverage limited through-plane context from MRI, which may constrain risk prediction performance.
In clinical interpretation, risk-related cues span multiple spatial scales, ranging from subtle short-range local lesion changes to long-range global patterns such as BPE and bilateral asymmetry~\cite{acciavatti2023beyond}.
Both types of cues are essential for accurate risk prediction with different temporal horizons, with local variations often reflecting short-term risk and global patterns indicating longer-term risk \cite{lauritzen2023assessing,wang2026incorporating}.

Inspired by prior multiple-instance learning (MIL) approaches~\cite{shazeer2017outrageously,ilse2018attention,shao2021transmil,yang2024mambamil,wang2020defense}, an MRI volume can be treated as a bag of instances (2D slices) associated with an exam-level label. 
Features extracted from individual slices by 2D CNNs are then aggregated by MIL to capture structural information of the 3D volume. 
The commonly used attention-based MIL (e.g., ABMIL~\cite{ilse2018attention}) approaches improve interpretability by learning instance importance weights.
However, such MIL frameworks provide limited modeling of long-range slice interactions, which is essential when global risk-related patterns (e.g., BPE and asymmetry) are distributed across many slices \cite{acciavatti2023beyond}. 
Transformer-based MIL variants such as TransMIL~\cite{shao2021transmil} introduce global self-attention, but their positional encodings are designed for 2D spatial tokens in whole-slide images rather than the 1D sequential order of MRI slices. 
Moreover, because slice features are extracted independently and aggregated as per-slice global descriptors, fine-grained local variations across adjacent slices are difficult to preserve. 
As a result, existing methods do not fully model both local and global cross-slice dependencies, potentially limiting short- and long-term BC risk prediction.

In this study, we propose \M, a 2.5D local–global omni-slice structural framework for efficient MRI-based 5-year BC risk prediction. 
\textbf{Our main contributions are as follows:}
\emph{1) We jointly model both short- and long-range cross-slice dependencies while maintaining 2D CNN efficiency.}
To align with clinical interpretation, the model first encodes local continuity via lightweight neighbor-slice stacking within 2D backbones to identify short-term risk cues. 
Subsequently, long-range slice interactions associated with long-term BC risk are captured by transformer-based MIL with MRI-aware positional encoding, producing interpretable slice importance.
\emph{2) To capture complementary anatomical structure across orientations, we extend the method to a multi-plane version, \MV, which models axial, sagittal, and coronal planes.} The multi-plane slice importance enables projection of exam-level risk to voxel-level saliency.
\emph{3) We conduct comprehensive validation on a large clinical screening cohort ($\sim$7.5K MRIs).} For both short- to long-term risk prediction, \M~consistently outperforms 3D CNNs and prior 2D-MIL methods while being substantially more efficient than 3D CNNs. The \MV~further improves predictive performance.

\section{Methods}
Given a breast MRI volume $V \in \mathbb{R}^{D \times H \times W}$, we treat it as an ordered sequence of axial slices $S = \{ s_{1}, s_{2}, \ldots, s_{D}\}$. 
Our proposed \M~ (Fig.~\ref{fig:method}~B) consists of two main modules: 
1) Local structural encoding, where neighboring slices are fused into a pseudo-RGB representation, followed by slice-level feature extraction using a 2D CNN; 
2) Global structural modeling is performed by a transformer-enhanced MIL aggregator that captures long-range slice relationships and produces interpretable slice-level importance scores. 
Moreover, \MV~ extends the same local–global modeling across the axial, sagittal, and coronal directions to enable complementary volumetric representation.

\textbf{Local cross-slice structural encoding (Lo):} 
To encode short-range inter-slice anatomical continuity, we augment each target slice with its neighboring slices to form a 2.5D input (i.e., a pseudo-3D slice representation \cite{vu2020evaluation}). Specifically, for slice $s_{i}$, we construct a pseudo-RGB representation by concatenating $s_{i-g}$, $s_{i}$, and $s_{i+g}$ along the channel dimension: 
$x_{i}= \mathrm{concat}(s_{i-g},\, s_{i},\, s_{i+g}) \in \mathbb{R} ^{3 \times H \times W}$, where $g$ denotes the slice gap. This three-channel design is well aligned with standard 2D CNN backbones and facilitates transfer learning from standard 2D vision backbones.
For boundary slices, we replicate the nearest valid slice.
This input formulation introduces local volumetric context while retaining the efficiency of 2D CNN architectures. A very small gap produces highly redundant neighbors, whereas a large gap may weaken local anatomical continuity. We therefore evaluate multiple gap sizes, $g \in \{0, 1, 3, 5, 7\}$, in ablation studies to quantify the effect of neighborhood context.
Each pseudo-RGB slice $x_{i}$ is then encoded independently by a 2D CNN backbone (e.g., ResNet18), 
$h_{i}= f_{\theta}(x_{i}) \in \mathbb{R}^{C}$, 
yielding an ordered sequence of slice embeddings $H = \{h_{1}, h_{2}, \ldots, h_{D}\}$. 
This design provides an efficient approximation of local volumetric continuity without adding trainable parameters \cite{vu2020evaluation}.

\textbf{Global structural modeling (Go):}
To model global structural dependencies across the ordered slice sequence, we employ an order-aware transformer and attention based MIL aggregator. First, given the slice embeddings ($H=\{h_1,h_2,\ldots,h_D\}$) from local encoding, we integrate slice-order information using a non-trainable continuous sinusoidal positional encoding \cite{wang2025predicting}. The positional encoding is broadcast to the slice sequence and added to $H$.
Then, the position-aware sequence is processed by a transformer encoder: $H'=\mathrm{Transformer}(H)\in\mathbb{R}^{D\times C}$. We use a lightweight transformer setting with 2 encoder layers and 8 attention heads to balance global dependency modeling and model complexity. Through the transformer, each slice representation is updated using information from all slices in the volume, enabling modeling of long-range inter-slice dependencies and global structural interactions beyond local neighborhood context. 
We subsequently apply attention-based MIL pooling \cite{ilse2018attention} for volume-level aggregation: $z_{\mathrm{bag}},\,\boldsymbol{\alpha}=\mathrm{AttnMIL}(H')$, where $z_{\mathrm{bag}}$ is the bag-level representation and $\boldsymbol{\alpha}$ denotes normalized slice-level importance scores. The pooling layer aggregates contextualized slice features into a volume representation, while the attention weights quantify each slice contribution to the volume-level risk prediction. This yields interpretable slice-level importance estimates and enables identification of risk-relevant slices. Finally, a linear prediction head maps $z_{\mathrm{bag}}$ to the final risk.

\textbf{Multi-plane \MV:}
To exploit complementary anatomical structure across orientations, \MV~extends \M~ to axial, coronal, and sagittal planes (Fig.~\ref{fig:method}~C). The input MRI volume is processed independently in the axial, coronal, and sagittal planes, and each plane is modeled by a separate \M~ network to produce a plane-specific risk score. The final risk prediction is computed by averaging the three plane-specific scores.
This late-fusion design enables orientation-specific representation learning while reducing cross-plane feature coupling, resulting in stable and robust multi-plane risk estimation. When attention-based MIL pooling is used, \MV also provides interpretable slice-level importance along each axis. These plane-wise importance weights can be combined to construct a three-dimensional risk saliency map: $ A(d,h,w) = w_z(d)\, w_y(h)\, w_x(w) $, which provides an approximate voxel-level attribution of risk-relevant regions within the breast volume.

\textbf{Risk formulation:}
Following prior BC risk prediction studies~\cite{wang2024ordinal,wang2025predicting}, all models predict an $(n+1)$-dimensional probability vector after sigmoid activation,
$\mathbf{p} = (p_1, \ldots, p_n, p_{n+1})$, where $p_t$ denotes the probability that BC is diagnosed in year $t$ after the index MRI, and $p_{n+1}$ denotes the probability of remaining healthy within $n$ years. 
For each exam, the ground truth is encoded as a binary vector $\mathbf{y} \in \{0,1\}^{n+1}$ with a single positive entry. If BC is diagnosed in year $t\in\{1,\ldots,n\}$, then $y_t=1$; if no BC is observed within the $n$-year prediction window, then $y_{n+1}=1$.
To handle right-censored follow-up, we use a time-dependent mask $\delta_t$ that excludes unobservable years beyond the available follow-up duration for healthy cases. The network is trained using a masked binary cross-entropy loss applied independently across years:
$\mathcal{L}=\frac{1}{\sum_t \delta_t}\sum_{t=1}^{n+1}\delta_t\left[-y_t\log p_t-(1-y_t)\log(1-p_t)\right]$,
where $\delta_t \in \{0,1\}$ indicates whether the outcome at year $t$ is observable given the follow-up duration. This loss ensures that learning is driven only by reliable supervision while properly handling censored samples.
Then, the cumulative risk scores of developing BC within $m$ years are computed as
$ \mathrm{Risk}_{\le m} = \sum_{t=1}^{m} p_t$,
allowing simultaneous estimation of one- to five-year risk from a single model output.

\begin{figure*}[t]
\centering
\begin{minipage}[b]{0.48\textwidth}
\centering
\includegraphics[width=\linewidth, height=0.09\textheight]{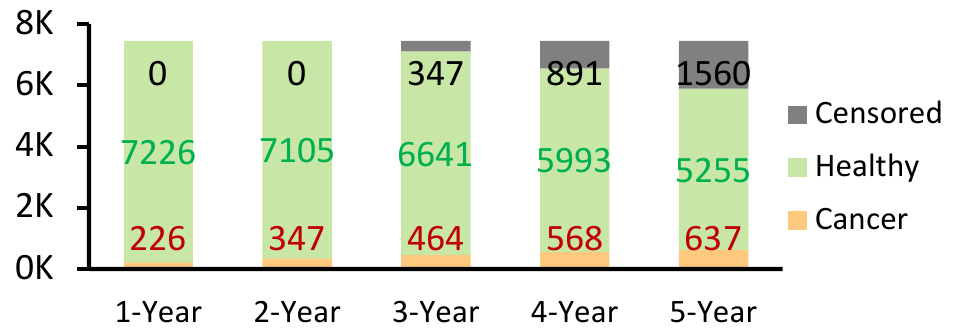}
\captionof{figure}{Distribution of censored, healthy, and cancer cases.
}
\label{fig:dataset_barplot}
\end{minipage}%
\hfill
\begin{minipage}[b]{0.49\textwidth}
\centering
\tiny
\setlength{\tabcolsep}{1pt}
\begin{tabular}{lcccc}
\bottomrule
Subset & Whole & Train & Val & Test\\
\hline
0--1 Year
& 226 (3.37\%)
& 112 (3.56\%)
& 66 (3.55\%)
& 48 (2.52\%) \\
1--2 Year
& 121 (2.21\%)
& 59 (2.53\%)
& 36 (1.94\%)
& 26 (1.37\%) \\
2--3 Year
& 117 (1.94\%)
& 54 (2.09\%)
& 34 (1.83\%)
& 29 (1.53\%) \\
3--4 Year
& 104 (1.69\%)
& 42 (1.72\%)
& 40 (2.15\%)
& 22 (1.16\%) \\
4--5 Year
& 69 (1.15\%)
& 24 (1.12\%)
& 30 (1.61\%)
& 15 (0.79\%) \\
\hline
Total MR & 7452 & 3692 & 1859 & 1901 \\
\bottomrule
\end{tabular}
\captionof{table}{Dataset composition by time-to-cancer interval for all splits.}
\label{tab:dataset_stats}
\end{minipage}
\end{figure*}

\section{Experiments}

\textbf{Dataset:}
The in-house dataset includes breast DCE-MRIs acquired between 2004 and 2020 with institutional review board approval. 
All exams are linked to longitudinal follow-up records indicating time to BC diagnosis or censoring. Data are split at the patient level into training, validation, and test sets with a ratio of 0.5/0.25/0.25 to prevent information leakage.
The distribution of time-to-cancer labels across splits is summarized in Table~\ref{tab:dataset_stats} and visualized in Fig.~\ref{fig:dataset_barplot}.

\textbf{Evaluation:}
Model performance is evaluated using the concordance index (C-index)~\cite{uno2011c} to assess overall risk ranking consistency under censoring. In addition, we report the AUC for one- to five-year risk prediction horizons, following prior BC risk modeling studies~\cite{wang2024ordinal,wang2025predicting}. For all performance metrics, 95\% confidence intervals were estimated using 1{,}000 bootstrap resamples of the test set and reported as mean $\pm$ 1.96 standard deviations~\cite{wang2024ordinal}. \emph{In this paper, all C-index and AUC metrics are reported as value$\times$100.} To assess efficiency, we report FLOPs and inference throughput (FPS) per MRI volume on the same hardware.

\textbf{Implementation:}
All MRI volumes are preprocessed following the method described in prior breast MRI studies \cite{gao2024explainable}. Volumes are resampled and cropped to a spatial resolution of $352\times192\times144$ while preserving the original aspect ratio. During training, whole volume-based data augmentations are applied includes random flipping, rotation, affine transformation, and translation.
All 2D backbones are initialized with ImageNet-pretrained weights, while 3D backbones are initialized with weights pretrained on large-scale 3D medical images \cite{chen2019med3d}.
Models are trained using the Adam optimizer with a learning rate of $5\times10^{-5}$ and a batch size of 1–4, depending on GPU memory. Early stopping is applied based on the validation C-index.

\begin{table}[!t]
\centering
\tiny
\centering
\setlength{\tabcolsep}{2pt}
\caption{Performance comparison of different methods across prediction
horizons. 
}
\label{tab:main_results}
\resizebox{\textwidth}{!}{
\begin{tabular}{llcc|cccccc c}
\toprule
& Method 
& FLOPs 
& FPS 
& 1Y AUC 
& 2Y AUC 
& 3Y AUC 
& 4Y AUC 
& 5Y AUC 
& Mean AUC 
& C-index \\

\midrule
\multirow{7}{*}{\rotatebox{90}{\textbf{ResNet18}}}

& 3D Baseline
& 1170 
& 4 
& \pct{0.568}\pmS{8.1}
& \pct{0.581}\pmS{6.6}
& \pct{0.567}\pmS{5.6}
& \pct{0.551}\pmS{5.1}
& \pct{0.556}\pmS{4.7}
& \pct{0.565}\pmS{5.4}
& \pct{0.564}\pmS{4.6} \\

& 3D MFFN\cite{wang2025accurate}
& 1187 
& 3 
& \pct{0.624}\pmS{7.5}
& \pct{0.606}\pmS{6.3}
& \pct{0.589}\pmS{5.5}
& \pct{0.568}\pmS{4.9}
& \pct{0.566}\pmS{4.7}
& \pct{0.591}\pmS{5.1}
& \pct{0.567}\pmS{4.6} \\

\cline{2-11}

& 2D Baseline (Mean)
& 354 
& 18 
& \pct{0.637}\pmS{9.3} 
& \pct{0.611}\pmS{7.2} 
& \pct{0.600}\pmS{5.7} 
& \pct{0.556}\pmS{5.3} 
& \pct{0.547}\pmS{5.1} 
& \pct{0.590}\pmS{5.9} 
& \pct{0.562}\pmS{4.9} \\

& 2D MoE~\cite{shazeer2017outrageously} 
& 354 
& 17 
& \pct{0.614}\pmS{7.9} 
& \pct{0.596}\pmS{6.7} 
& \pct{0.579}\pmS{5.6} 
& \pct{0.590}\pmS{5.0} 
& \pct{0.599}\pmS{4.8} 
& \pct{0.596}\pmS{5.3} 
& \pct{0.576}\pmS{4.6} \\

& 2D LSTM-MIL~\cite{wang2020defense} 
& 354 
& 17 
& \pct{0.678}\pmS{8.0} 
& \pct{0.646}\pmS{6.5} 
& \pct{0.616}\pmS{5.5} 
& \pct{0.585}\pmS{5.1} 
& \pct{0.575}\pmS{4.8} 
& \pct{0.620}\pmS{5.4} 
& \pct{0.588}\pmS{4.7} \\

& 2D MambaMIL~\cite{yang2024mambamil} 
& 354 
& 17 
& \pct{0.683}\pmS{7.7} 
& \pct{0.602}\pmS{6.8} 
& \pct{0.580}\pmS{5.9} 
& \pct{0.573}\pmS{5.2} 
& \pct{0.568}\pmS{5.0} 
& \pct{0.601}\pmS{5.4} 
& \pct{0.562}\pmS{4.8} \\

& 2D ABMIL~\cite{ilse2018attention} 
& 354 
& 18 
& \pct{0.715}\pmS{7.6} 
& \pct{0.685}\pmS{6.1} 
& \pct{0.647}\pmS{5.5} 
& \pct{0.635}\pmS{5.0} 
& \pct{0.624}\pmS{4.8} 
& \pct{0.661}\pmS{5.0} 
& \pct{0.619}\pmS{4.7} \\

& 2D TransMIL~\cite{shao2021transmil} 
& 354 
& 17 
& \pct{0.687}\pmS{8.7} 
& \pct{0.635}\pmS{6.9} 
& \pct{0.615}\pmS{5.8} 
& \pct{0.603}\pmS{5.3} 
& \pct{0.604}\pmS{5.0} 
& \pct{0.629}\pmS{5.6} 
& \pct{0.594}\pmS{4.9} \\

\rowcolor{C1}
& \textcolor{black}{\M-RISK} & 354 & 18 
& \textcolor{red}{\pct{0.771}\pmS{6.7}} 
& \underline{\textcolor{black}{\pct{0.696}\pmS{6.7}}} 
& \underline{\textcolor{black}{\pct{0.658}\pmS{5.5}}} 
& \underline{\textcolor{black}{\pct{0.639}\pmS{5.0}}} 
& \underline{\textcolor{black}{\pct{0.633}\pmS{4.8}}} 
& \underline{\textcolor{black}{\pct{0.679}\pmS{5.2}}} 
& \textbf{\textcolor{red}{\pct{0.631}\pmS{4.8}}} \\

\rowcolor{C2}
& \textcolor{black}{\MV-RISK} 
& 1082 & 6 & \underline{\textcolor{black}{\pct{0.754}\pmS{7.2}}} 
& \textbf{\textcolor{red}{\pct{0.697}\pmS{6.0}}} 
& \textbf{\textcolor{red}{\pct{0.675}\pmS{5.1}}} 
& \textbf{\textcolor{red}{\pct{0.659}\pmS{4.9}}}
& \textbf{\textcolor{red}{\pct{0.650}\pmS{4.7}}} 
& \textbf{\textcolor{red}{\pct{0.687}\pmS{4.8}}} 
& \underline{\textcolor{black}{\pct{0.630}\pmS{4.6}}} \\

\midrule
\multirow{7}{*}{\rotatebox{90}{\textbf{ResNet50}}}

& 3D Baseline
& 1518 
& 2 
& \pct{0.600}\pmS{8.7}
& \pct{0.576}\pmS{7.0}
& \pct{0.574}\pmS{6.0}
& \pct{0.583}\pmS{5.5}
& \pct{0.585}\pmS{5.1}
& \pct{0.584}\pmS{5.8}
& \pct{0.559}\pmS{4.9} \\

& 3D MFFN \cite{wang2025accurate}
& 1575 
& 2 
& \pct{0.677}\pmS{5.0}
& \pct{0.640}\pmS{8.8}
& \pct{0.625}\pmS{7.1}
& \pct{0.615}\pmS{6.0}
& \pct{0.613}\pmS{5.3}
& \pct{0.634}\pmS{5.2}
& \pct{0.613}\pmS{5.8} \\

\cline{2-11}

& 2D Baseline (Mean)
& 801 
& 5 
& \pct{0.616}\pmS{8.1} 
& \pct{0.612}\pmS{6.0} 
& \pct{0.598}\pmS{5.3} 
& \pct{0.578}\pmS{5.0} 
& \pct{0.575}\pmS{4.8} 
& \pct{0.596}\pmS{5.2} 
& \pct{0.574}\pmS{4.6} \\

& 2D MoE~\cite{shazeer2017outrageously} 
& 803 
& 5 
& \pct{0.708}\pmS{7.7} 
& \pct{0.658}\pmS{6.2} 
& \pct{0.642}\pmS{5.4} 
& \pct{0.634}\pmS{5.0} 
& \pct{0.634}\pmS{4.7} 
& \pct{0.655}\pmS{5.1} 
& \pct{0.611}\pmS{4.6} \\

& 2D LSTM-MIL~\cite{wang2020defense} 
& 810 
& 5 
& \pct{0.714}\pmS{7.8} 
& \pct{0.652}\pmS{6.6} 
& \pct{0.627}\pmS{5.6} 
& \pct{0.620}\pmS{5.2} 
& \pct{0.621}\pmS{5.0} 
& \pct{0.646}\pmS{5.2} 
& \pct{0.598}\pmS{4.8} \\

& 2D MambaMIL~\cite{yang2024mambamil} 
& 809 
& 5 
& \pct{0.681}\pmS{8.8} 
& \pct{0.629}\pmS{6.9} 
& \pct{0.596}\pmS{6.1} 
& \pct{0.577}\pmS{5.6} 
& \pct{0.587}\pmS{5.2} 
& \pct{0.614}\pmS{5.8} 
& \pct{0.577}\pmS{5.2} 
\\

& 2D ABMIL~\cite{ilse2018attention} 
& 802 
& 5 
& \underline{\pct{0.724}\pmS{8.7}} 
& \pct{0.651}\pmS{7.1} 
& \pct{0.625}\pmS{5.9} 
& \pct{0.606}\pmS{5.3} 
& \pct{0.606}\pmS{5.0} 
& \pct{0.642}\pmS{5.7} 
& \pct{0.589}\pmS{5.0} \\

& 2D TransMIL~\cite{shao2021transmil} 
& 802 
& 5 
& \pct{0.709}\pmS{6.8} 
& \pct{0.637}\pmS{6.0} 
& \pct{0.615}\pmS{5.3} 
& \pct{0.621}\pmS{4.8} 
& \textbf{\textcolor{red}{\pct{0.638}\pmS{4.4}}}
& \pct{0.644}\pmS{4.8} 
& \pct{0.603}\pmS{4.4} \\

\rowcolor{C1}

& \textcolor{black}{\M-RISK} 
& 811 
& 5 
& \textcolor{black}{\pct{0.721}\pmS{8.2}} 
& \underline{\textcolor{black}{\pct{0.680}\pmS{6.6}}} 
& \underline{\textcolor{black}{\pct{0.654}\pmS{5.5}}} 
& \underline{\textcolor{black}{\pct{0.639}\pmS{4.9}}} 
& \underline{\textcolor{black}{\pct{0.631}\pmS{4.7}}} 
& \underline{\textcolor{black}{\pct{0.665}\pmS{5.4}}} 
& \underline{\textcolor{black}{\pct{0.625}\pmS{4.7}}} \\

\rowcolor{C2}

& \textcolor{black}{\MV-RISK} 
& 2482 
& 2 
& \textbf{\textcolor{red}{\pct{0.729}\pmS{8.0}}}
& \textbf{\textcolor{red}{\pct{0.691}\pmS{6.5}}} 
& \textbf{\textcolor{red}{\pct{0.662}\pmS{5.2}}} 
& \textbf{\textcolor{red}{\pct{0.639}\pmS{4.8}}} 
& \textcolor{black}{\pct{0.627}\pmS{4.7}} 
& \textbf{\textcolor{red}{\pct{0.670}\pmS{5.2}}} 
& \textbf{\textcolor{red}{\pct{0.629}\pmS{4.6}}} \\

\bottomrule
\end{tabular}
}
\end{table}

\begin{figure}[!t]
\center
\includegraphics[width=\textwidth]{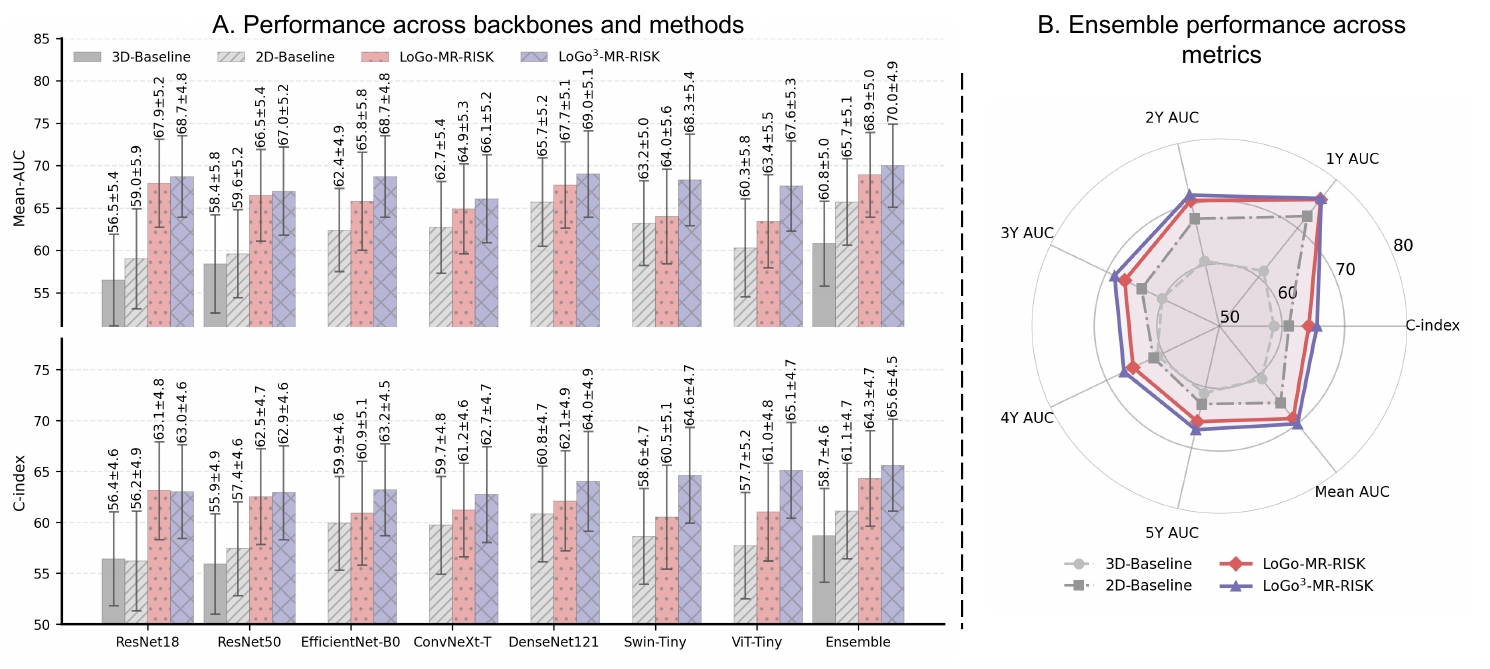}
\caption{Comparison of 3D CNNs, 2D baselines, and the proposed \M-RISK and \MV-RISK across seven backbones.}\label{fig:result_bar}
\end{figure}

\section{Results and Discussion}
\textbf{Comparison experiments:} 
Table~\ref{tab:main_results} compares \M~with 3D baseline, the 3D multi-scale feature fusion network (MFFN)~\cite{wang2025accurate}, and representative 2D SOTA MIL approaches. Across both ResNet-18 and -50 backbones, \M~consistently ranks among the top methods in C-index and AUCs for one- to five-year risk prediction.
In particular, \M~improves the C-index by $\sim$5–7\% over the 2D mean-pooling baseline and by a substantial margin over the fully 3D baseline. In addition, \M~reduces computational cost relative to the fully 3D CNNs, with lower FLOPs and FPS. 
These results indicate that explicitly encoding inter-slice anatomical structure through architectural design can be more effective than increasing model dimensionality alone, especially in data-constrained clinical settings. 
Compared with other MIL-based 2D models, including MoE \cite{shazeer2017outrageously}, LSTM-MIL \cite{wang2020defense}, ABMIL \cite{ilse2018attention}, TransMIL \cite{shao2021transmil}, and MambaMIL \cite{yang2024mambamil}, \M~yields better overall performance across short- to long-term risk prediction, suggesting improved modeling of spatially extended risk patterns in MRI volumes. Moreover, the multi-plane extension \MV~slightly improves performance over single-plane \M.

\begin{figure}[!t]
\center
\includegraphics[width=\textwidth]{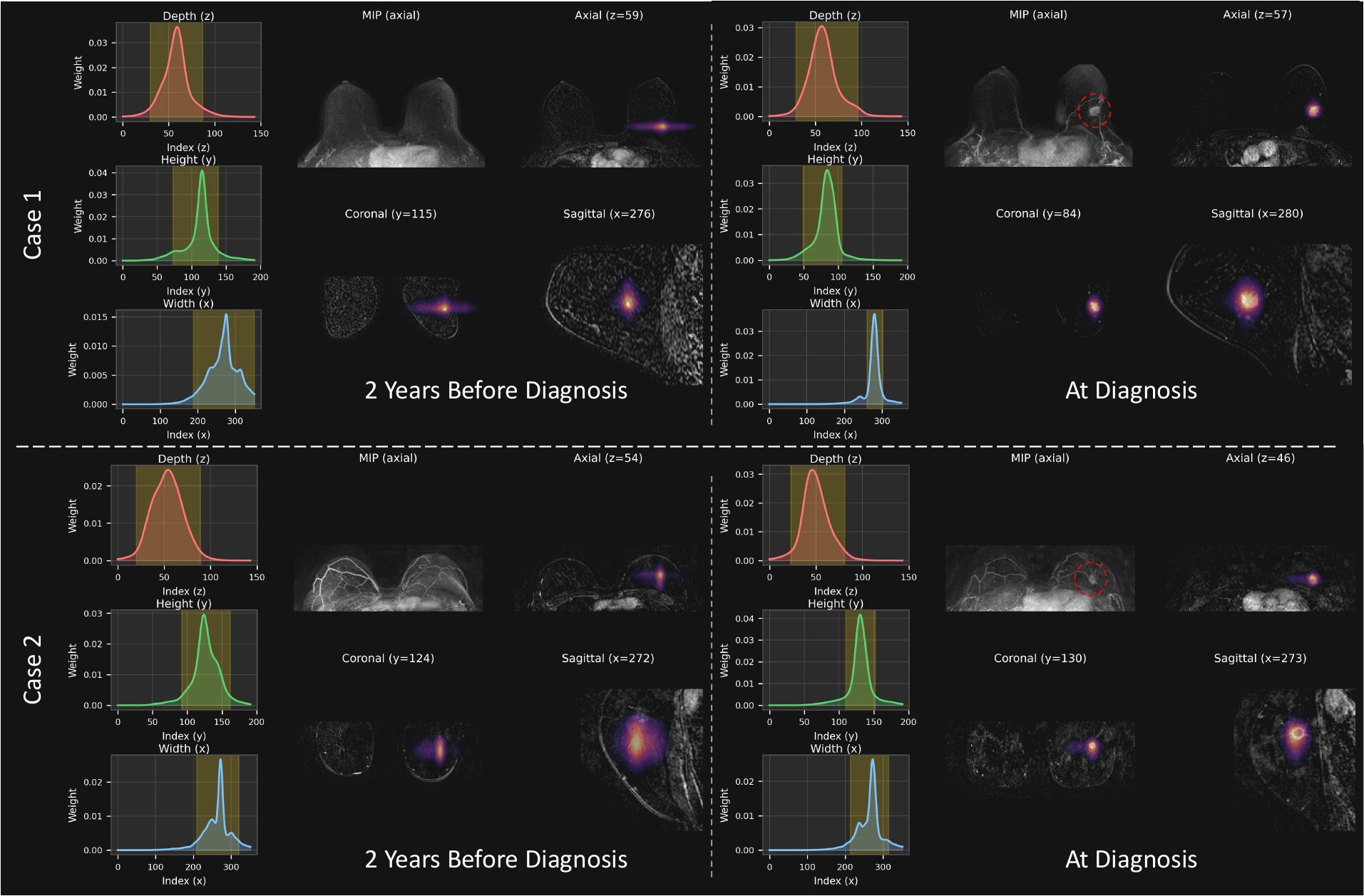}
\caption{Multi-plane risk localization with \MV. For each case, we present:
(Left) Slice-level importance weights learned along the axial (Z), coronal (Y), and sagittal (X) planes. (Right) The maximum intensity projection (MIP) view and corresponding high-importance regions onto orthogonal views, illustrating spatially consistent risk-relevant areas localization across planes over time.
}
\label{fig:result_vis}
\end{figure}

\begin{table}[!t]
\centering
\tiny
\setlength{\tabcolsep}{2pt}
\caption{Ablation study of local and global structural modeling. 
}
\label{tab:ablation}
\resizebox{\textwidth}{!}{
\begin{tabular}{llccclcccccc}
\toprule
& Lo 
& Go 
& Method 
& 1Y AUC 
& 2Y AUC 
& 3Y AUC 
& 4Y AUC 
& 5Y AUC 
& Mean AUC 
& C-index \\

\midrule
\multirow{9}{*}{\rotatebox{90}{\textbf{ResNet18}}}
& \textcolor{gray}{$\times$} 
& \textcolor{gray}{$\times$} 
& 2D Baseline: $g=0$
& \pct{0.637}\pmS{9.3} 
& \pct{0.611}\pmS{7.2} 
& \pct{0.600}\pmS{5.7} 
& \pct{0.556}\pmS{5.3} 
& \pct{0.547}\pmS{5.1} 
& \pct{0.590}\pmS{5.9} 
& \pct{0.562}\pmS{4.9} \\

\cline{2-11}

& \textcolor{red}{\checkmark} 
& \textcolor{gray}{$\times$} 
& Gap: $g=1$ 
& \pct{0.660}\pmS{9.1} 
& \pct{0.595}\pmS{7.1} 
& \pct{0.558}\pmS{6.1} 
& \pct{0.533}\pmS{5.6} 
& \pct{0.529}\pmS{5.2} 
& \pct{0.575}\pmS{5.9} 
& \pct{0.551}\pmS{5.0} \\

& \textcolor{red}{\checkmark} 
& \textcolor{gray}{$\times$} 
& Gap: $g=3$
& \pct{0.689}\pmS{8.2} 
& \pct{0.632}\pmS{6.6} 
& \pct{0.590}\pmS{5.6} 
& \pct{0.568}\pmS{5.1} 
& \pct{0.562}\pmS{4.9} 
& \pct{0.608}\pmS{5.4} 
& \pct{0.575}\pmS{4.7} \\

& \textbf{\textcolor{red}{\checkmark}} 
& \textbf{\textcolor{gray}{$\times$}} 
& \textcolor{orange}{Gap: $g=5$}
& \textcolor{orange}{\pct{0.678}\pmS{8.4}}
& \textcolor{orange}{\pct{0.630}\pmS{6.8}}
& \textcolor{orange}{\pct{0.606}\pmS{5.6}} 
& \textcolor{orange}{\pct{0.583}\pmS{5.0}} 
& \textcolor{orange}{\pct{0.590}\pmS{4.8}}
& \textcolor{orange}{\pct{0.618}\pmS{5.4}} 
& \textcolor{orange}{\pct{0.592}\pmS{4.7}} \\

& \textcolor{red}{\checkmark} 
& \textcolor{gray}{$\times$} 
& Gap: $g=7$
& \pct{0.639}\pmS{8.8} 
& \pct{0.620}\pmS{6.9} 
& \pct{0.593}\pmS{5.8} 
& \pct{0.561}\pmS{5.2} 
& \pct{0.547}\pmS{5.0} 
& \pct{0.592}\pmS{5.6} 
& \pct{0.568}\pmS{4.8} \\

\cline{2-11}

& \textcolor{gray}{$\times$} 
& \textcolor{red}{\checkmark} 
& w/o Lo \& w/o Pos
& \pct{0.649}\pmS{9.0} 
& \pct{0.602}\pmS{7.4} 
& \pct{0.598}\pmS{6.0} 
& \pct{0.587}\pmS{5.3} 
& \pct{0.587}\pmS{5.0} 
& \pct{0.605}\pmS{5.9} 
& \pct{0.583}\pmS{5.0} \\

& \textcolor{gray}{$\times$} 
& \textcolor{red}{\checkmark} 
& \textcolor{orange}{w/o Lo} 
& \textcolor{orange}{\pct{0.726}\pmS{7.4}} 
& \textcolor{orange}{\pct{0.688}\pmS{6.2}} 
& \textcolor{orange}{\pct{0.648}\pmS{5.4}} 
& \textcolor{orange}{\pct{0.632}\pmS{5.1}} 
& \textcolor{orange}{\pct{0.632}\pmS{4.9}} 
& \textcolor{orange}{\pct{0.665}\pmS{5.0}} 
& \textcolor{orange}{\pct{0.620}\pmS{4.8}} \\

\rowcolor{C1}
& \textcolor{red}{\checkmark} 
& \textcolor{red}{\checkmark} 
& \textcolor{black}{\M-RISK}
& \textcolor{black}{\pct{0.771}\pmS{6.7}} 
& \textcolor{black}{\pct{0.696}\pmS{6.7}} 
& \textcolor{black}{\pct{0.658}\pmS{5.5}} 
& \textcolor{black}{\pct{0.639}\pmS{5.0}} 
& \textcolor{black}{\pct{0.633}\pmS{4.8}} 
& \textcolor{black}{\pct{0.679}\pmS{5.2}} 
& \textcolor{black}{\pct{0.631}\pmS{4.8}} \\

\rowcolor{C2}
& \textcolor{red}{\checkmark} 
& \textcolor{red}{\checkmark} 
& \textcolor{black}{\MV-RISK}
& \textcolor{black}{\pct{0.754}\pmS{7.2}} 
& \textcolor{black}{\pct{0.697}\pmS{6.0}} 
& \textcolor{black}{\pct{0.675}\pmS{5.1}} 
& \textcolor{black}{\pct{0.659}\pmS{4.9}} 
& \textcolor{black}{\pct{0.650}\pmS{4.7}} 
& \textcolor{black}{\pct{0.687}\pmS{4.8}} 
& \textcolor{black}{\pct{0.630}\pmS{4.6}} \\

\bottomrule
\end{tabular}
}
\end{table}

Fig.~\ref{fig:result_bar} summarizes model performance across seven backbones.
\M~and \MV~ consistently outperform the 2D and 3D baselines, indicating that the gains are mainly from the proposed local–global structural modeling rather than backbone choice. 
The framework may further benefit from advances in 2D feature extractors and adapt to different computational constraints without architectural modification. 
Moreover, we leverage an ensemble strategy that combines predictions from multiple backbones. 
The ensemble \MV~achieves the best performance across metrics (Fig.~\ref{fig:result_bar}B), suggesting complementary representations across backbones and improved robustness. 

\textbf{Visualization analysis:}
Fig.~\ref{fig:result_vis} presents multi-plane risk localization examples from \MV, including exams acquired 2 years before diagnosis and at diagnosis. The model produces slice-level importance weights along three directions, and projects the corresponding high-importance regions onto orthogonal MRI views. In both cases, the multi-plane local-global modeling captures coherent volumetric risk patterns. Notably, high-risk regions in the pre-diagnosis exams are consistent with the tumor regions at the time of diagnosis. These visualizations support the interpretability of \MV~and suggest that the learned multi-plane importance can provide meaningful cues for risk localization.

\textbf{Ablation studies:}
Table~\ref{tab:ablation} analyzes the contribution of individual components in \M. 
For the local module, using neighbor-slice fusion with an intermediate slice gap improves performance over the 2D baseline, while too small or too large gaps are less effective. In particular, $g=5$ provides the best overall trade-off across horizons and is therefore used in the final model. For the global module, transformer-based MIL aggregation improves risk prediction over the baseline setting, and adding positional encoding further improves both AUC and C-index, indicating the importance of preserving slice order in sequence modeling. The full \M~configuration, which jointly combines local cross-slice encoding and global sequence-level aggregation, achieves the best overall performance, supporting the complementary roles of short-range anatomical continuity and long-range slice dependency modeling.

\section{Conclusion}
We presented \M, a 2.5D local-global inter-slice structural modeling framework for breast MRI-based cancer risk prediction. It combines local neighbor-slice encoding and global transformer-based MIL aggregation, for effective volumetric structure modeling. Experiments on a large institutional screening cohort show that \M~ consistently outperforms 2D/3D baselines and MIL-based methods across seven backbones. The multi-plane extension \MV~ further improves risk prediction and provides interpretable slice-level importance across axial, sagittal, and coronal planes, offering meaningful cues for risk localization without requiring a fully 3D architecture. Overall, these results support explicit inter-slice structural modeling as an effective and efficient strategy for breast MRI risk stratification, and position \M/\MV~ as a practical foundation for future multi-modal BC risk prediction.

%
\bibliographystyle{unsrt}
\bibliography{ref.bib}

\end{document}